\documentclass[11pt]{article}

\usepackage[preprint]{acl}

\usepackage{times}
\usepackage{latexsym}

\usepackage[T1]{fontenc}
\usepackage[utf8]{inputenc}

\usepackage{microtype}

\usepackage{inconsolata}

\usepackage{graphicx}

\usepackage{booktabs}
\usepackage{fvextra}

\title{Elias in the Lighthouse, Again? Diagnosing Low Diversity in LLM Stories}

\author{Sil Hamilton \and David Mimno \\
  Department of Information Science \\
  Cornell University \\
  \texttt{\{srh255,mimno\}@cornell.edu} \\}

\begin{document}
\maketitle
\begin{abstract}
LLM-generated stories are a popular use case, but they show very low variability.
We sample 20,000 total stories from four current models using five prompts. 
We find that 11 words occur in 88.3\% of generated stories, with little difference between models.
These words include names (Elias, Mara, Elara), settings (lighthouses), and professions (clockmaker, librarian).
These tokens do not often occur in published literature nor pre-training data, but they are found in preference data that is likely to have been used by all current models.
Surprisingly, these ``lighthouse'' stories are infrequent when compared with the average post-training story, much of which contains references to copyrighted characters or adult content. 
This result demonstrates the potentially disproportionate impact of small datasets combined with powerful alignment algorithms.
\end{abstract}

\section{Introduction}
\begin{figure}[t]
    \centering
    \includegraphics[width=0.485\textwidth]{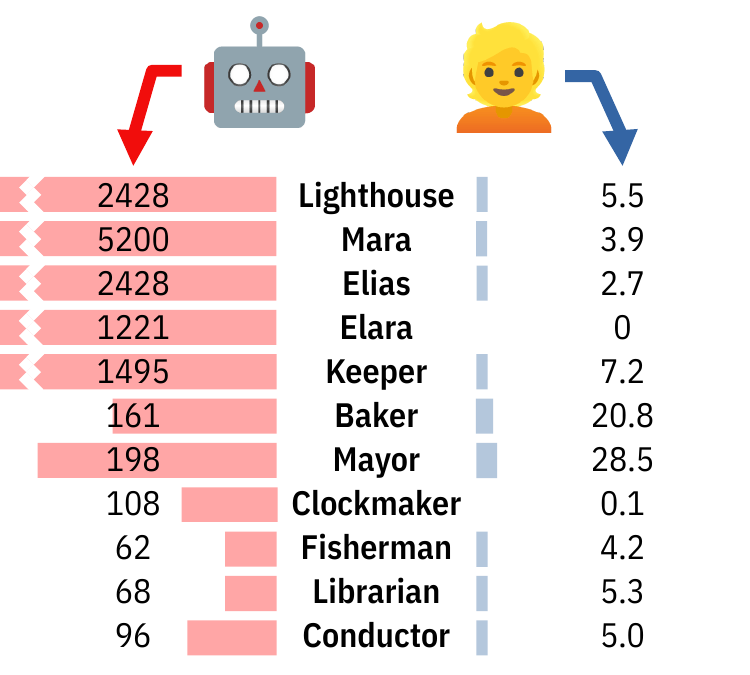}
    \caption{We prompt four models to write 20{,}000 stories --- 88.3\% of these stories contain at least one of 11 tokens at rates vastly higher than contemporary English literature (here measured in parts per million).}
    \label{fig:figure1}
\end{figure}
The output of large language models (LLMs), even across model families, is becoming increasingly homogeneous.
This mode-collapse phenomenon is unusually clear in creative writing \citep{hamiltonDetectingModeCollapse}.
While we know from prompt data that story-writing is a popular use case \cite{zhaoWildChat1MChatGPT2024}, and that readers prefer interesting and surprising literature \citep{morettiSlaughterhouseLiterature2000}, the stories generated by LLMs are remarkable in their sameness.

Prior work has proposed post-hoc solutions like adjusting sampling techniques \citep{troshinControlTemperatureSelective2025} and new post-training optimization objectives \citep{chungModifyingLargeLanguage2025}.
In this short paper we characterize story mode collapse and explore publicly available training data to locate the source.

We generate 20,000 stories with four current models from OpenAI, Anthropic, Google, and the Allen Institute for AI (AI2), finding 88.3\% of generated stories contain one of 11 core words (including character names, story locations, and professions). Most notably, over half feature a lighthouse.
Why does this story pattern become favoured?

These 11 words are not common in published English literature, which suggests that post-training data is responsible (\autoref{fig:figure1}).
But examining OLMo 3's post-training set reveals only 3,053 out of a total 78,958 stories contain one of our 11 words.\footnote{We release the IDs of these documents \href{https://github.com/srhm-ca/elias/}{here}.}
We find that the dominance of ``Elias in the Lighthouse'' stories cannot be explained by prevalence in pre- or post-training data.
We speculate that models are trained to avoid references to copyrighted characters and adult content during alignment but defer this question to future work.

\section{Related Work}
Mode collapse, the tendency for a generative model to overfit on a small set of samples during training, was first observed in LLMs after post-training techniques like SFT and RLHF were introduced \citep{ouyangTrainingLanguageModels2022,baiTrainingHelpfulHarmless2022,hamiltonDetectingModeCollapse}.
Fears of mode collapse deepened with growing synthetic data in training \citep{gerstgrasserModelCollapseInevitable2024,shumailovAIModelsCollapse2024}.
Mode collapse has been demonstrated in linguistic markers \citep{paechSampaechSlopforensics2025}, query answers \citep{zhangNoveltyBenchEvaluatingLanguage2025a,lagzianMultiNoveltyImproveDiversity2025}, and stories \citep{doshiGenerativeAIEnhances2024}.
Prior work has proposed modifying prompts \citep{lagzianMultiNoveltyImproveDiversity2025}, sampling techniques \citep{troshinControlTemperatureSelective2025}, and post-training algorithms \citep{kirkUnderstandingEffectsRLHF2024} --- but the phenomenon remains poorly understood.
To our knowledge, this study is the first to study a specific instance of mode collapse by tracing output back to training input.

\section{Characterizing Repetitive Story Traits}

\begin{figure}[t!]
{\setlength{\fboxsep}{8pt}%
\setlength{\fboxrule}{0.5pt}%
\fcolorbox{gray!40}{gray!10}{%
  \begin{minipage}{\dimexpr\columnwidth-2\fboxsep-2\fboxrule\relax}
    \itshape
    \setlength{\parskip}{0.6em}
    \setlength{\parindent}{0pt}
    The lighthouse at the edge of the world did not guide ships; it signaled to the stars.

    Elias had been the keeper for forty years. He was a man composed of salt, solitude, and the rhythmic ticking of gears. The lighthouse was a towering spire of obsidian, carved directly into a jagged needle of rock that rose from a sea so still it looked like polished slate.

    Every night, Elias climbed the two hundred and twelve steps to the lantern room. He didn't use oil or electricity. Instead, he tended to a sphere of captured nebula --- a swirling, violet bruise of light that pulsed in time with his own heartbeat\ldots
  \end{minipage}%
}}
\caption{Written by Gemini 3.1 Flash-Lite when prompted to ``write a story.'' Lighthouses are present in half of all 20{,}000 stories generated for this experiment.}
\label{fig:lighthouse}
\end{figure}

Current LLMs from OpenAI, Anthropic, Google, and AI2 write similar stories when prompted to ``write a story'' with no additional constraints.\footnote{We find similar behavior for more complex prompts, but we focus on simple prompts for this initial study.}
We use four current models: Claude Haiku 4.5, Gemini 3.1 Flash-Lite, GPT-5.4-Mini, and OLMo 7b Thinking.\footnote{We selected smaller models to maximize sample size for a fixed budget, we observe large and small models belonging to the same family share storywriting behavior.}
We prompt each model with five requests (``Write a story,'' ``Please write a story,'' ``Write me a story,'' ``Tell me a story,'' and ``Please tell a story'') 1,000 times each, yielding 20,000 total stories totalling 12.8 million words.\footnote{All models were accessed via OpenRouter for a total cost of \$180 U.S. dollars. Endpoints available as of April 2026.}

A typical example shown in \autoref{fig:lighthouse} highlights three elements common across nearly all 20,000 stories: a location (19,864 stories), a character name (19,864 stories), and a profession (15,807 stories).
In fact, the specific location (``lighthouse''), name (``Elias''), and profession (``keeper'') in this story appear in some combination across 66.6\% of all generated stories.
\textit{Light} is likewise a common theme: 56\% of stories generated by Claude are titled ``The Lighthouse Keeper’s Secret'' and the word ``light'' appears in 16,784 stories at an average rate of 3.2 instances per story. %(cf. \autoref{tab:light-word}).

Other common names include Mara and Elara; locations include lighthouses and villages; and professions include clockmakers, fishermen, and librarians.
Nearly all stories combine two or more of these three elements, suggesting models are sampling each from some common pool of candidates. What other words does the pool contain?

To construct vocabulary lists useful for downstream analysis, we use GPT-5.4-nano to identify token spans corresponding with story settings, characters' first names, and their professions.\footnote{All prompts are in \autoref{appendix:prompts}.}
We verify the presence of each extracted span before filtering candidates in three steps: (i) we tokenize strings on whitespace, yielding multiple tokens per string; (ii) for each story and category we retain the extracted token with the highest corpus-level frequency; and (iii) we retain all tokens emitted by at least half of all models.
Removing incoherent candidates yields 663 tokens: 247 locations, 71 names, and 345 professions across all stories.

\begin{table*}[t]
  \begin{minipage}[t]{0.555\textwidth}
  \vspace{1em}
  \centering
  \small
  \begin{tabular}{p{1.3cm}r|rrrrrr}
    \toprule
     & & & \textsc{PRE-} & \textsc{PRE-} & \textsc{POST-} & \textsc{POST-}\\
     & \textsc{Ours} & \textsc{Lit} & \textsc{NON} & \textsc{FIC} & \textsc{NON} & \textsc{FIC}\\
    \midrule
    \multicolumn{6}{l}{\textbf{Name}} \\
    \midrule
    ``elias'' & 2,428 & 2.7 & 2.2 & 4.0 & 0.4 & \textbf{52.7} \\
    ``mara'' & 5,200 & 3.9  & 2.5 & 8.7 & 0.4 & \textbf{21.7} \\
    ``elara'' & 1,221 & 0.0 & 0.4 & 1.2 & 0.9 & \textbf{108} \\
    \midrule
    \multicolumn{3}{l}{\textbf{Profession}}\\
    \midrule
    ``keeper'' & 1,495 & 7.2 & 6.3 & \textbf{14.7} & 3.5 & 10.0 \\
    ``baker'' & 161 & 20 & 11.8 & 10.56 & 1.7 & \textbf{11.9} \\
    ``mayor'' & 198 & \textbf{28} & 11.5 & 16.1 & 1.4 & 27.4 \\
    ``clockmaker'' & 108 & 0.1 & 0.18 & 0.0 & 0.3 & \textbf{1.4} \\
    ``fisherman'' & 62 & 4.2 & 3.0 & 7.6 & 0.0 & \textbf{9.3} \\
    ``librarian'' & 68 & 5.3 & 7.6 & 5.9 & 2.3 & \textbf{11.5} \\
    ``conductor'' & 96 & 5.0 & 5.9 & 5.7 & 4.7 & \textbf{7.5} \\
    % ``elias'' & 2428.7 & 2.7 & 2.8 & 0.6 & 0.4 & \textbf{52.7} \\
    % ``mara'' & 5200.7 & 3.9 & 4.5 & 0.4 & 0.4 & \textbf{21.7} \\
    % ``elara'' & 1221.5 & 0.0 & 0.7 & 1.3 & 0.9 & \textbf{108.6} \\
    % \midrule
    % \multicolumn{3}{l}{\textbf{Profession}}\\
    % \midrule
    % ``keeper'' & 1495.1 & 7.2 & 9.0 & 3.6 & 3.5 & \textbf{10.0} \\
    % ``baker'' & 161.7 & 20.8 & 11.4 & 1.8 & 1.7 & \textbf{11.9} \\
    % ``mayor'' & 198.0 & \textbf{28.5} & 13.0 & 1.5 & 1.4 & 27.4 \\
    % ``clockmaker'' & 108.4 & 0.1 & 0.2 & 0.0 & 0.0 & \textbf{1.4} \\
    % ``fisherman'' & 62.5 & 4.2 & 4.5 & 0.4 & 0.0 & \textbf{9.3} \\
    % ``librarian'' & 68.7 & 5.3 & 7.0 & 10.4 & 2.3 & \textbf{11.5} \\
    % ``conductor'' & 96.0 & 5.0 & 5.9 & 4.7 & 4.7 & \textbf{7.5} \\
    \midrule
    \multicolumn{3}{l}{\textbf{Location}}\\
    \midrule
    ``lighthouse'' & 3,005 & 5.5 & 3.5 & 4.6 & 4.6 & \textbf{10.1} \\
    % ``lighthouse'' & 3005.7 & 5.5 & 4.7 & 4.6 & 4.6 & \textbf{10.1} \\
    \bottomrule
  \end{tabular}
  \caption{Counts for the most frequent words in our corpus as measured in parts per million words (PPM) versus representative samples of English literature (\textsc{LIT}), (non-)fiction web data (\textsc{PRE-NON/FIC}), and (non-)fiction post-training data (\textsc{POST-NON/FIC}). Corpora containing the most Core tokens in bold.}
  \label{tab:conlit-ratios}
  \end{minipage}
  \hfill
  \begin{minipage}[t]{0.38\textwidth}
  \vspace{0pt}
  \centering
  \small
  \begin{tabular}{lrrr}
    \toprule
    Stage & Stories & Any Core & Core\% \\
    \midrule
    SFT & 68{,}674 & 2{,}092 & 3.0\% \\
    DPO & 6{,}876 & 548 & 8.0\% \\
    RL & 3{,}408 & 370 & 10.9\% \\
    \bottomrule
  \end{tabular}
  \caption{Core rates by alignment stage.}
  \label{tab:t1-by-stage}
  \vspace{1em}
  \begin{tabular}{lrr}
    \toprule
    Source & Stories & Core\% \\
    \midrule
    WildChat-derived & 59{,}276 & 2.6\% \\
    \midrule
    DPO & 4{,}990 & 8.1\% \\
    \texttt{persona-precise-if-r1} & 3{,}988 & 6.9\% \\
    \texttt{allenai} & 3{,}222 & 6.9\% \\
    \texttt{IF\_multi\_constraints} & 1{,}650 & 20.4\% \\
    \texttt{if\_qwq\_reasoning} & 1{,}689 & 5.4\% \\
    \texttt{rlvr\_general\_mix} & 1{,}751 & 1.9\% \\
    \texttt{ultrafeedback} & 341 & 13.8\% \\
    \texttt{wildguardmix-r1} & 101 & 22.8\% \\
    \texttt{aya-100k-r1} & 895 & 1.8\% \\
    \texttt{other} & 1{,}055 & 4.2\% \\
    \bottomrule
  \end{tabular}
  \caption{Core story prevalence by source. Alignment datasets have 5--8x higher Core density than WildChat-derived stories despite being 80\% of all post-training stories.}
  \label{tab:t1-by-source}
  \end{minipage}
\end{table*}

Within the candidate vocabulary words, we additionally select a Core vocabulary of 11 words using a changepoint analysis to find a minimal set of candidate tokens most common across all stories \citep{killick2012optimal}.
88.3\% of stories contain a Core token.
The Core includes names (``Elias'' is in 26.5\% of all stories, ``Mara'' in 16.7\%, and ``Elara'' in 13.1\%), professions (``keeper'' at 48.1\% of all stories, while ``clockmaker,'' ``baker,'' ``fisherman,'' ``librarian,'' ``mayor,'' and ``conductor'' each occur in 1.9\% to 6.6\% of stories), and a single location: ``lighthouse'' with a frequency of 51.2\%.

The core words and a second tier of 50 additional words are given in \autoref{appendix:common}, while Core PPM rates per models are shown in \autoref{tab:t1-ppm-by-model}.
98\% of stories contain at least one of these 61 words, while 49.1\% contain a full name-profession-location triple.
Words (especially names) vary by model as shown in \autoref{appendix:by-model}, but nearly all terms are used by all models.
Professions suggest an idyllic, pre-modern setting: clockmaker, blacksmith, innkeeper, keeper, baker, fisherman. Other tokens describe curation (restorer, collector, caretaker).

\section{Tracing Story Traits To Training Data}
\begin{figure*}[t]
    \centering
    \includegraphics[width=\textwidth]{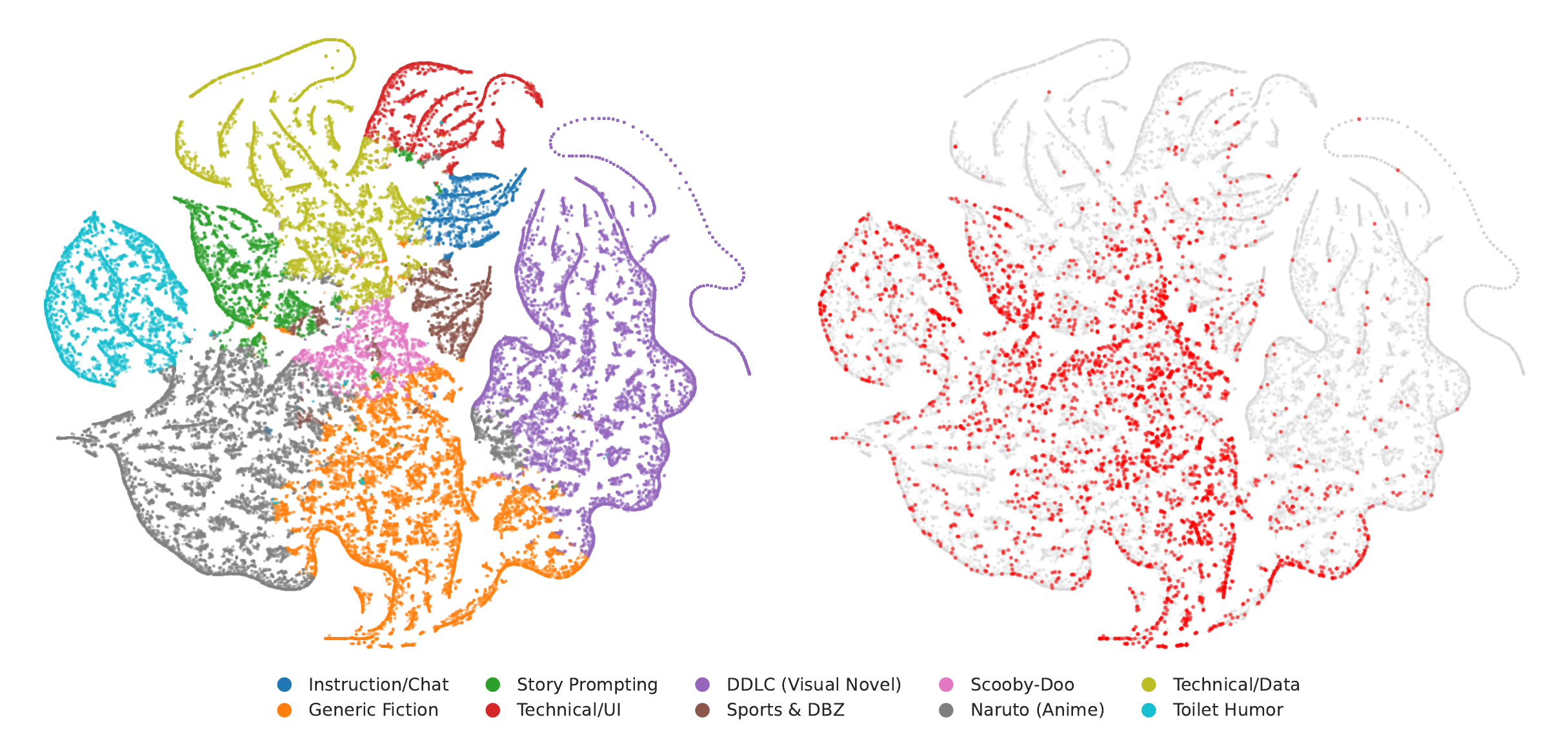}
    \caption{t-SNE of Topic model over all stories in OLMo 3's post-training set (left), and Core stories (right). Stories highlighted in red are ``lighthouse'' stories, spread across many topics, including toilet humor and fan fiction.}
    \label{fig:tsne}
\end{figure*}
Frequent Core vocabulary in LLM story generation cannot be explained by the frequency of those words in published English fiction, pre-training data, or post-training data.
We assess each potential source by comparing Core rates in our corpus with English corpora.
We give rates in \autoref{tab:conlit-ratios}.

The simplest explanation would be that Core tokens are common in English literature.
We consulted CONLIT, a corpus containing 2,700 contemporary English novels published between 2007-2021 across 12 genres of fiction with $\approx287$ million total words \citep{piperCONLITDatasetContemporary2022}.
The frequency of Core tokens is far greater in generated stories than published fiction, e.g. ``Elias'' is 900 times more frequent in our corpus.
To assess amateur fiction we consulted stories on the subreddit \texttt{/r/writingprompts} \citep{huang2024gpt}.
Rates are similar to CONLIT (\autoref{appendix:reddit}), suggesting models do not track human storywriting patterns.

To assess whether these tokens are common in English web data, we turn to OLMo 3, whose training data includes Common Crawl and is freely available.
OLMo 3 was trained on $\approx$ 3.89 billion predominantly human written documents during pre-training, of which 33 million are marked as \textit{Literature}.
Across these documents we find near-negligible Core PPM rates (e.g. ``Elara'' appears 0.7 times per million words).
To ensure that we are looking at web \textit{stories} and not non-fiction literature,
we train a fiction classifier with 200k balanced samples from OLMo's pre-training corpus annotated for narrativity with GPT-OSS 20b with the following prompt inspired by \citet{piper2025narradetect}: ``Is this passage a work of fiction? Answer only with a number: 1 if yes, 0 if no.''
We then train a FastText classifier and evaluate on 400 balanced samples of fiction and non-fiction in CONLIT, achieving a $F_1$ of 0.84 (precision = 0.75, recall = 0.98).
Filtering by this classification shows a slight ($\approx 2 \times$) increase in some Core words in the \textit{fiction} portion, but nowhere near the rate in our generated stories.

If Core words are not common in web data, then one remaining source would be post-training data.
But we find that OLMo's post-training data exhibits our tokens at a \textit{lower} rate than CONLIT.
Using the same fiction classifier, we find that 78{,}958 \textit{stories} in post-training data show the highest concentration of Core words of any of the training or literature subsets, but even then dramatically less than generated stories: ``Elias'' occurs 52.7 PPM in OLMo 3 stories vs. 2.7 in CONLIT, but 2428 in our corpus.

\paragraph{Which dataset(s) are contributing Core tokens?}
This suggests OLMo 3 learns to write Core stories from relatively few samples.
To understand which datasets are contributing these stories, we assign a binary score to each story indicating the presence of one or more Core tokens.\footnote{For documents containing accepted/rejected pairs, we only consider the accepted sample.}
We expected the majority of Core stories to appear in SFT data because WildChat (and derivatives) are the most story dominant source for OLMo at 59{,}266 total stories \citep{zhaoWildChat1MChatGPT2024}.
But only 1,803 of these stories contain Core tokens, and measuring the Core rate by alignment stage (e.g. SFT, DPO, and RL) shows DPO \& RL contribute relatively more Core stories than SFT (\autoref{tab:t1-by-stage} \& \autoref{tab:t1-by-source}).
We find OLMo 3 learns Core vocabulary from 3,053 examples, or 3.8\% of all stories observed during post-training.

\section{Post-Training Story Genres}
To better understand what kinds of stories OLMo 3 encounters during post-training, we trained a 10-topic LDA topic model \citep{blei2003latent} against the full post-training story corpus (\autoref{fig:tsne}).
We find a diverse range of content, with dominant topics including fan fiction for popular Japanese media, video games, and American cartoons.
As expected given their relative frequency, ``lighthouse'' stories do not form a single topic, and are instead spread out across our discovered topics.
They are particularly concentrated in clusters containing generic fiction, but they nonetheless fail to dominate any topic.
A close reading reveals several topics frequently feature stories containing inappropriate humour and adult content, surprising considering OLMo 3 will not typically emit inappropriate content when writing.
Future work will want to investigate whether these stories fail to trigger safety and quality filters used for data cleaning, and why if so.

\section{Conclusion}
When given little direction, current frontier models write stories using a narrow catalog of names, places, and occupations.
Recurring characters in these stories include Elias, a lighthouse keeper.
Elias is unusual; the name is uncommon in literature, web data, and even post-training data.
We have found that of the 78,958 stories exposed to OLMo 3 during post-training, only $\approx 3{,}053$ stories contain one or more of these 11 unusual tokens.
But despite constituting only 3.8\% of post-training stories --- and $7.71 \times 10^{-7}$ of the $\approx4$ billion total documents OLMo 3 was trained on --- these ``lighthouse'' stories hold a disproportionately large influence over what stories the model writes in practice.
This suggests models do not simply mimic the dominant patterns in their training corpora.
Future work will want to determine whether alignment causes models to prefer the ``safest'' (for work) samples in post-training, avoiding the potentially unsafe topic matter of many stories they otherwise encounter.

\section*{Limitations}
Our experiment is expressly monolingual to prevent confounders stemming from unanticipated multilingual behaviour, but it would be valuable for future work to explore how multilingual storywriting prompts impact the phenomena observed.

\bibliography{custom}

\appendix

\section{Prompts}
\label{appendix:prompts}
In this section we provide all prompts used over the course of the experiment.

\paragraph{Prompts for story generation.}
The five prompts used to generate stories are as follows.

\texttt{Write a story.}

\texttt{Please write a story.}

\texttt{Write me a story.}

\texttt{Tell me a story.}

\texttt{Please tell a story.}

\paragraph{Name, location, profession extraction.}
The system prompt for extracting names, locations, and professions is as follows.

{\linespread{0.92}\selectfont
\begin{Verbatim}[breaklines=true,
                 breakautoindent=true,
                 breakindent=2mm,
                 xleftmargin=2mm,
                 xrightmargin=0mm,
                 formatcom=\color{black!75}]
You extract structured metadata from short stories.

Return JSON only, with this exact schema:
{
  "character_names": ["first name only", ...],
  "settings": ["place or location noun phrase", ...],
  "professions": ["profession or role noun", ...]
}

Rules:
- Output valid JSON and nothing else.
- `character_names` must contain only first names for named human or human-like characters.
- Exclude surnames, titles, pronouns, groups, and unnamed roles.
- `settings` should be concise setting/location phrases from the story, e.g. "Lighthouse", "Village square", "Oakhaven".
- `professions` should be concise occupations or stable roles, e.g. "Clockmaker", "Lighthouse keeper", "Baker".
- Deduplicate while preserving first appearance order.
- If a field has no items, return an empty list.
- Every returned string must be an exact token span copied from the story text.
\end{Verbatim}

The user prompt for extracting names, locations, and professions is as follows.

{\linespread{0.92}\selectfont
\begin{Verbatim}[breaklines=true,
                 breakautoindent=true,
                 breakindent=2mm,
                 xleftmargin=2mm,
                 xrightmargin=0mm,
                 formatcom=\color{black!75}]
Read the story and answer these questions:
1. Who are the characters in the text?
2. What is the character's role in the text?
3. What is the setting?

Return JSON only in this exact schema:
{{
  "character_names": ["first name only", ...],
  "settings": ["place or location noun phrase", ...],
  "professions": ["profession or stable role noun phrase", ...]
}}

Additional rules:
- For `character_names`, include only first names for named human or human-like characters.
- For `professions`, map each character's role in the text to concise occupations or stable roles when present.
- For `settings`, list the main setting locations or place names.
- Every returned item must be an exact contiguous span copied from the story text.
- Do not normalize, paraphrase, singularize, pluralize, or invent labels.
- If the exact answer is not present as a span in the story, omit it.
- Deduplicate items while preserving first appearance order.
- Use empty lists when something is absent.

Story:
{story}
\end{Verbatim}

The prompt for identifying fictionality is as follows.

{\linespread{0.92}\selectfont
\begin{Verbatim}[breaklines=true,
                 breakautoindent=true,
                 breakindent=2mm,
                 xleftmargin=2mm,
                 xrightmargin=0mm,
                 formatcom=\color{black!75}]
``Is this passage a work of fiction? Answer only with a number: 1 if yes, 0 if no.''
\end{Verbatim}

\section{Common Tokens}
In \autoref{tab:npl-tokens} we present all 61 Core and additional tokens common to LLM generated short stories we identified in our experiment.
\label{appendix:common}
\begin{table*}[t]
\centering
\small
\setlength{\tabcolsep}{4pt}
\begin{minipage}[t]{0.22\textwidth}
\centering
\textbf{Names}\\[2pt]
\begin{tabular}{lr}
\toprule
Token & Count \\
\midrule
\multicolumn{2}{l}{\textit{Core}} \\
elias  & 5{,}294 \\
mara   & 3{,}345 \\
elara  & 2{,}627 \\
\midrule
\multicolumn{2}{l}{\textit{Additional}} \\
marcus & 1{,}822 \\
thomas & 1{,}263 \\
lila   & 1{,}262 \\
clara  & 1{,}260 \\
sarah  & 685 \\
maya   & 633 \\
elian  & 420 \\
mira   & 418 \\
\bottomrule
\end{tabular}
\end{minipage}\hfill
\begin{minipage}[t]{0.50\textwidth}
\centering
\textbf{Professions}\\[2pt]
\begin{tabular}{lr@{\hskip 14pt}lr}
\toprule
Token & Count & Token & Count \\
\midrule
\multicolumn{4}{l}{\textit{Core}} \\
keeper      & 9{,}609 & fisherman    & 673 \\
baker       & 1{,}325 & librarian    & 592 \\
mayor       & 975     & conductor    & 389 \\
clockmaker  & 958     &              &     \\
\midrule
\multicolumn{4}{l}{\textit{Additional}} \\
guard       & 1{,}454 & apothecary   & 100 \\
captain     & 1{,}417 & curator      & 94  \\
guardian    & 1{,}037 & mapmaker     & 93  \\
sailor      & 514     & watchmaker   & 86  \\
jeweler     & 383     & weaver       & 84  \\
priest      & 352     & historian    & 80  \\
owner       & 287     & innkeeper    & 80  \\
collector   & 281     & scholar      & 80  \\
blacksmith  & 247     & clerk        & 73  \\
miller      & 244     & hermit       & 64  \\
elder       & 201     & farmer       & 64  \\
doctor      & 198     & custodian    & 61  \\
restorer    & 193     & engineer     & 47  \\
cartographer& 155     & lawyer       & 45  \\
archivist   & 151     & stationmaster& 44  \\
caretaker   & 143     & scientist    & 35  \\
apprentice  & 126     & mender       & 32  \\
shopkeeper  & 123     & healer       & 27  \\
teacher     & 122     & scavenger    & 21  \\
postman     & 101     & proprietor   & 17  \\
\bottomrule
\end{tabular}
\end{minipage}\hfill
\begin{minipage}[t]{0.22\textwidth}
\centering
\textbf{Locations}\\[2pt]
\begin{tabular}{lr}
\toprule
Token & Count \\
\midrule
\multicolumn{2}{l}{\textit{Core}} \\
lighthouse & 10{,}233 \\
\midrule
\multicolumn{2}{l}{\textit{Additional}} \\
village & 4{,}155 \\
attic   & 1{,}978 \\
\bottomrule
\end{tabular}
\end{minipage}
\caption{Tokens in the first two changepoint segments of each category's coverage curve, with token hit counts across the generated story corpus. Within each segment tokens are listed in descending hit-count order.}
\label{tab:npl-tokens}
\end{table*}

\section{Per-Model Core PPM}
\label{appendix:by-model}
We present Core token concentrations across all stories generated by each model in \autoref{tab:t1-ppm-by-model}.

\begin{table*}[t]
  \centering
  \small
  \setlength{\tabcolsep}{10pt}
  \begin{tabular}{l l rrrrr}
    \toprule
    & & \multicolumn{1}{c}{Claude} & \multicolumn{1}{c}{Gemini} & \multicolumn{1}{c}{GPT} & \multicolumn{1}{c}{OLMo} & \\
    Token & Cat. & \multicolumn{1}{c}{Haiku 4.5} & \multicolumn{1}{c}{Flash-Lite} & \multicolumn{1}{c}{5.4-Mini} & \multicolumn{1}{c}{7B Think} & \multicolumn{1}{c}{All} \\
    \midrule
    \multicolumn{7}{l}{\textbf{Name}} \\
    \midrule
    elias & N & 9.3 & \textbf{10,752.4} & 899.0 & 893.2 & 2,483.4 \\
    mara & N & 47.5 & 0.5 & \textbf{10,718.0} & 84.8 & 5,317.8 \\
    elara & N & 2.6 & 153.1 & 110.2 & \textbf{6,625.5} & 1,249.0 \\
    \midrule
    \multicolumn{7}{l}{\textbf{Profession}} \\
    \midrule
    keeper & P & \textbf{4,185.4} & 1,776.9 & 899.0 & 713.5 & 1,528.7 \\
    baker & P & 0.0 & 35.6 & \textbf{298.6} & 65.8 & 165.4 \\
    mayor & P & 0.0 & 3.7 & \textbf{311.6} & 271.7 & 202.5 \\
    clockmaker & P & 0.0 & \textbf{258.2} & 123.3 & 27.2 & 110.9 \\
    fisherman & P & 51.6 & 11.4 & 77.3 & \textbf{88.9} & 63.9 \\
    librarian & P & 18.1 & 15.1 & 22.6 & \textbf{304.4} & 70.3 \\
    conductor & P & 112.5 & 11.9 & \textbf{158.9} & 0.9 & 98.1 \\
    \midrule
    \multicolumn{7}{l}{\textbf{Location}} \\
    \midrule
    lighthouse & L & \textbf{9,011.8} & 3,868.2 & 1,328.3 & 1,958.7 & 3,073.4 \\
    \bottomrule
  \end{tabular}
  \caption{PPM for each Core token across all models. Bolded values indicate each token's most frequent model.}
  \label{tab:t1-ppm-by-model}
\end{table*}

\section{Core Token Frequencies on Reddit}
\label{appendix:reddit}
We calculate Core token frequencies on non-published human-written literature from Reddit in \autoref{tab:wp-core}.

\begin{table}[t]
  \centering
  \small
  \begin{tabular}{lrrrr}
    \toprule
    Token & Count & PPM & Stories & \% \\
    \midrule
    \multicolumn{5}{l}{\textbf{Name}} \\
    \midrule
    ``elias''  & 230 & 1.25  & 50    & 0.02\% \\
    ``mara''   & 518 & 2.82  & 110   & 0.04\% \\
    ``elara''  & 6   & 0.03  & 3     & 0.00\% \\
    \midrule
    \multicolumn{5}{l}{\textbf{Profession}} \\
    \midrule
    ``keeper''     & 1{,}189 & 6.47  & 768   & 0.28\% \\
    ``baker''      & 1{,}145 & 6.23  & 596   & 0.22\% \\
    ``mayor''      & 2{,}343 & 12.74 & 1{,}073 & 0.39\% \\
    ``clockmaker'' & 24      & 0.13  & 15    & 0.01\% \\
    ``fisherman''  & 516     & 2.81  & 308   & 0.11\% \\
    ``librarian''  & 826     & 4.49  & 428   & 0.16\% \\
    ``conductor''  & 462     & 2.51  & 304   & 0.11\% \\
    \midrule
    \multicolumn{5}{l}{\textbf{Location}} \\
    \midrule
    ``lighthouse'' & 543 & 2.95 & 235 & 0.09\% \\
    \midrule
    \textsc{Any Core} & 7{,}802 & 42.42 & --- & --- \\
    \bottomrule
  \end{tabular}
  \caption{Core token frequencies in the human-written WritingPrompts corpus which contains 272{,}600 stories.}
  \label{tab:wp-core}
\end{table}

\end{document}